\documentclass[a4paper,fleqn]{cas-dc}
\usepackage{amssymb}
\usepackage{hyperref}
\usepackage{graphicx}
\usepackage[caption=false]{subfig}
\usepackage{lipsum}
\usepackage{lscape}
\usepackage{amsmath}
\usepackage{multirow}
\usepackage[figuresright]{rotating}
\usepackage{lineno}
\usepackage{algorithm} 
\usepackage{algpseudocode} 
\usepackage{lineno}
\usepackage{nomencl}
\makenomenclature
\usepackage{calc}
\usepackage[T1]{fontenc}
\usepackage{relsize}

\usepackage{setspace}

\usepackage[sort,comma,authoryear,round]{natbib}

\setcitestyle{authoryear,open={(},close={)},citesep={;}} 

\hypersetup{
  colorlinks,
  citecolor=Violet,
  linkcolor=Red,
  urlcolor=Blue}

\newcolumntype{P}[1]{>{\centering\arraybackslash}p{#1}}

\def\tsc#1{\csdef{#1}{\textsc{\lowercase{#1}}\xspace}}
\tsc{WGM}
\tsc{QE}
\tsc{EP}
\tsc{PMS}
\tsc{BEC}
\tsc{DE}

\begin{document}
\let\WriteBookmarks\relax
\def\floatpagepagefraction{1}
\def\textpagefraction{.001}
\shorttitle{Foundation Models in Smart Agriculture: Basics, Opportunities, and Challenges}
\shortauthors{Li et~al.}

\title [mode = title]{Large Language Models and Foundation Models in Smart Agriculture: Basics, Opportunities, and Challenges}

\author[1]{Jiajia Li}\ead{lijiajia@msu.edu}
\author[2]{Mingle Xu}\ead{xml@jbnu.ac.kr}
\author[3]{Lirong Xiang}\ead{lxiang3@ncsu.edu}
\author[1]{Dong Chen}\ead{chendon9@msu.edu}
\author[4]{Weichao Zhuang}\ead{wezhuang@seu.edu.cn}
\author[5]{Xunyuan Yin}\ead{xunyuan.yin@ntu.edu.sg}
\author[6]{Zhaojian Li}\ead{lizhaoj1@egr.msu.edu}

\address[1]{Department of Electrical and Computer Engineering, Michigan State University, East Lansing, MI, USA}
\address[2]{Department of Electronic Engineering, Core Research Institute of Intelligent Robots, Jeonbuk National University, South Korea}
\address[3]{Department of Biological and Agricultural Engineering, North Carolina State University, Raleigh, NC, USA}
\address[4]{School of Mechanical Engineering, Southeast University, Nanjing, China}
\address[5]{School of Chemistry, Chemical Engineering and Biotechnology, Nanyang Technological University, Singapore}
\address[6]{Department of Mechanical Engineering, Michigan State University, East Lansing, MI, USA}
\address{* Zhaojian Li and Xunyuan Yin are the corresponding authors.}

\begin{abstract}
The past decade has witnessed the rapid development and adoption of machine and deep learning (ML \& DL) methodologies in agricultural systems, showcased by great successes in applications such as smart crop management, smart plant breeding, smart livestock farming, precision aquaculture farming, and agricultural robotics. However, these conventional ML/DL models have certain limitations: they heavily rely on large, costly-to-acquire labeled datasets for training, require specialized expertise for development and maintenance, and are mostly tailored for specific tasks, thus lacking generalizability. Recently, large pre-trained models, also known as foundation models (FMs), have demonstrated remarkable successes in language, vision, and decision-making tasks across various domains. These models are trained on a vast amount of data from multiple domains and modalities. Once trained, they can accomplish versatile tasks with just minor fine-tuning and minimal task-specific labeled data. Despite their proven effectiveness and huge potential, there has been little exploration of applying FMs to agriculture artificial intelligence (AI). Therefore, this study aims to explore the potential of FMs in the field of smart agriculture. In particular, conceptual tools and technical background are presented to facilitate the understanding of the problem space and uncover new research directions in this field. To this end, recent FMs in the general computer science (CS) domain are reviewed, and the models are categorized into four categories: language FMs, vision FMs, multimodal FMs, and reinforcement learning FMs. Subsequently, the process of developing agriculture FMs (AFMs) is outlined and their potential applications in smart agriculture are discussed. In addition, the unique challenges and risks associated with developing AFMs are discussed, including model training, validation, and deployment. Through this study, the advancement of AI in agriculture is explored by introducing AFMs as a promising paradigm that can significantly mitigate the reliance on extensive labeled datasets and enhance the efficiency, effectiveness, and generalization of agricultural AI systems. To facilitate further research, a well-classified and actively updated list of papers on AFMs is organized and accessible at \url{https://github.com/JiajiaLi04/Agriculture-Foundation-Models}.
\end{abstract}


\maketitle

\section{Introduction}
\label{sec:intro}

\begin{figure*}[!t]
    \centering
{\includegraphics[width=.88\linewidth]{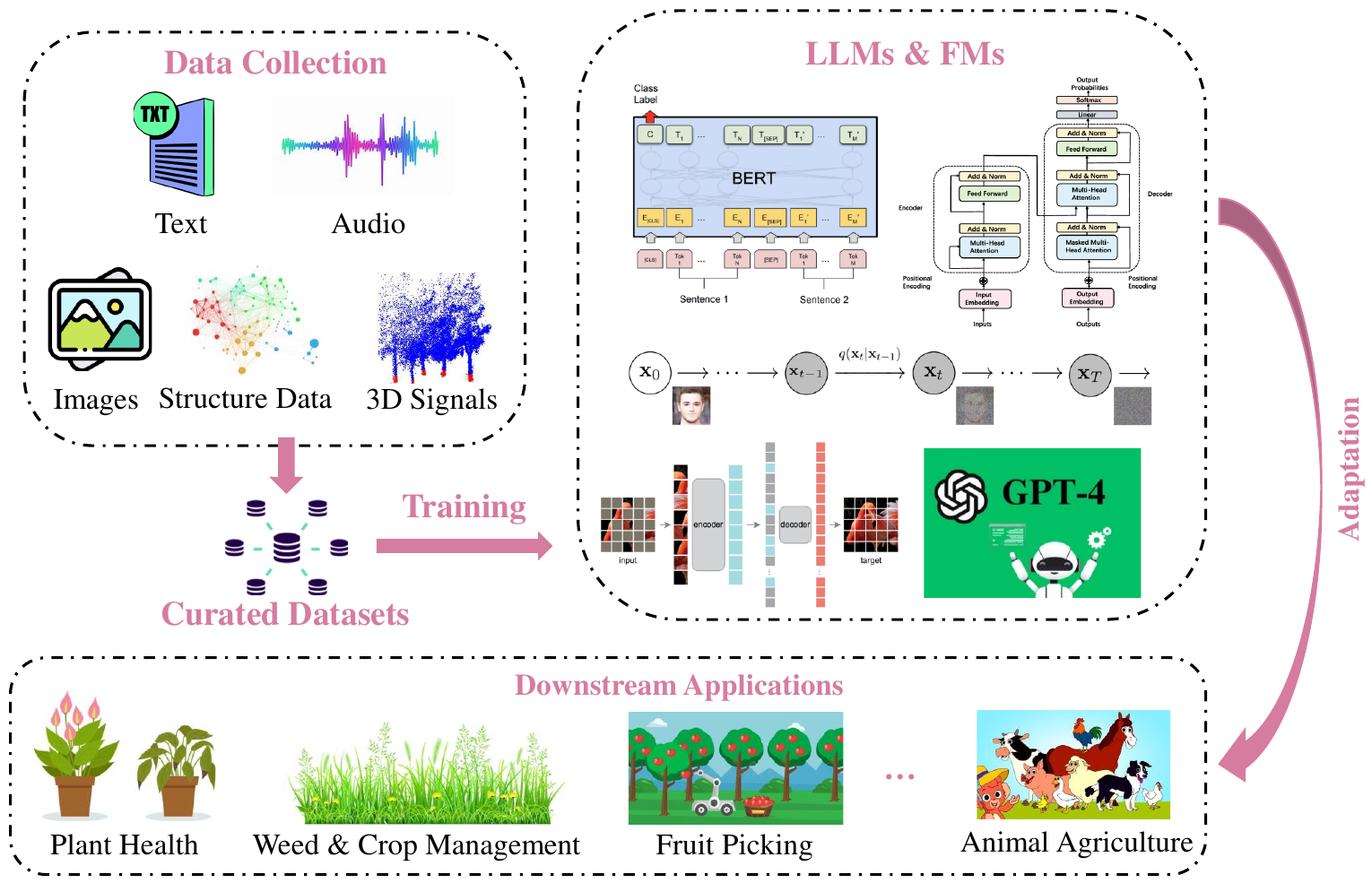} }
    \caption{An overview of building FMs for agricultural applications. The pipeline depicts the process of (multi-modal) data collection and cleaning, training or fine-tuning of LLMs and FMs, and downstream adaptation. The terms ``BERT'' and ``GPT-4'' refer to Bidirectional Encoder Representations from Transformers (BERT) \citep{devlin2018bert} and Generative Pre-trained Transformer 4 (GPT-4) \citep{achiam2023gpt}, respectively.}
    \label{fig:pipeline}
    \vspace{-10pt}
\end{figure*}

Modern agricultural systems have been undergoing a revolution through the tight integration of recent advances in information and communication technologies (ICT). This has led to a prosperous field known as smart farming (also referred to as smart agriculture) \citep{gondchawar2016iot, wolfert2017big, walter2017smart}, showing promise in enhancing agricultural outputs, increasing farming efficiency, and improving the quality of the final product. These technologies feature multi-disciplinary advancements involving unmanned aerial/ground vehicles (UAVs/UGVs), image processing, machine learning, big data, cloud computing, and wireless sensor networks (WSNs) \citep{walter2017smart,wolfert2017big, moysiadis2021smart}, enabling farmers to make well-informed decisions regarding planting, tending, and harvesting to maximize productivity and profits. However, efficient extraction of relevant information from diverse data sources, especially imaging data, presents a significant challenge. Traditional data mining approaches often struggle to uncover meaningful insights from such complex data \citep{wolfert2017big}. For instance, in agriculture, these techniques may face difficulties in extracting relevant features from datasets comprising diverse variables such as soil composition, weather patterns, crop health metrics, and geographical information \citep{wolfert2017big}.

Deep learning (DL), on the other hand, has demonstrated remarkable capabilities in processing complex and high-dimensional data due to its ability to learn hierarchical representations of data \citep{lecun2015deep}. Unlike traditional data mining techniques, which often rely on handcrafted feature extraction methods, DL models can automatically learn relevant features from raw data through multiple layers of abstraction. DL methods excel in feature extraction, pattern classification, and learning high-quality image representations, showing promising performance across various agricultural domains \citep{kamilaris2018deep}. These applications include weed control \citep{hasan2021survey, chen2022performance, li2024performance, chen2024synthetic}, plant disease detection \citep{mohanty2016using, xu2021style}, plant counting \citep{pathak2022review, li2023soybeannet}, plant phenotyping \citep{li2020review, xiang2023review}, postharvest quality assessment \citep{mendigoria2021vision, zhou2022deep}, robotic fruit harvesting \citep{chu2021deep, zhang2022algorithm, chu2023o2rnet}, among others. Despite the progress, these approaches heavily rely on supervised training, which hinges on large-scale, task-specific, and high-quality labeled datasets \citep{sun2017revisiting}. However, collecting and annotating such datasets is extremely time-consuming, resource-intensive, and expensive, thereby, creating a high bar for resource-limited applications \citep{lu2020survey, xu2023embracing}.  This challenge is particularly pronounced in specific applications such as plant disease detection, weed recognition, and fruit defect detection, where constraints related to biological materials, imaging conditions, and the need for precise annotations \citep{lu2020survey}. Moreover, the collected datasets are often limited in their generalizability to even similar agricultural domains or applications, necessitating repetition in the data collection and model development process \citep{ghazi2017plant, thenmozhi2019crop}. This repetitive process not only adds to the overall time and monetary costs but also hinders the generalizability, efficiency, and scalability of DL frameworks in agricultural applications.

\begin{table*}[!ht]
\renewcommand{\arraystretch}{1.4}
\centering
\caption{Nomenclature}
\label{tab:nomen1}
\resizebox{0.98 \textwidth}{!}{%
\begin{tabular}{|ll|ll|}
\hline
\textbf{Nomenclature} &                                                                                                 & InstructGPT & Instruction generative pretrained transformer \citep{ouyang2022training}                                                                  \\ 
ML           & Machine learning                                                                                & SAM         & Segment anything model \citep{kirillov2023segment}                                                                                        \\ 
DL           & Deep learning                                                                                   & SAA+        & Segment any anomaly + \citep{cao2023segment}                                                                                              \\ 
AI           & Artificial intelligence                                                                        & SEEM        & Segment everything everywhere model \citep{zou2023segment}                                                                                \\ 
GAN          & Generative adversarial network                                                                  & DMs         & Diffusion models                                                                                                                                           \\ 
VQA          & Visual question answering                                                                       & CLIP        & Contrastive language-image pre-training \citep{radford2021learning}                                                                       \\ 
FMs          & Foundation models                                                                               & ViT         & Vision transformer                                                                                                                                         \\ 
LFMs         & Language foundation models                                                                      & DALL$\cdot$E2     & Hierarchical text-conditional image generation with CLIP latent \citep{ramesh2022hierarchical} \\ 
VFMs         & Vision foundation models                                                                         & GLIDE       & Guided language to image diffusion for generation and editing \citep{nichol2021glide}                                                     \\ 
MFMs         & Multimodal foundation models                                                                     & GigaGAN     & Large-scale GAN for text-to-image synthesis \citep{kang2023scaling}                                                                       \\ 
RLFMs        & Reinforcement learning foundation models                                                        & BLIP        & Bootstrapping language-image pre-training \citep{li2022blip}                                                                              \\ 
BERT         & Bidirectional encoder representations from transformers \citep{devlin2018bert} & VLP         & Vision-language pre-training                                                                                                                               \\ 
PaLM         & Pathways language model \citep{wei2022chain}                                   & KOSMOS-1    & A multimodal large language model \citep{huang2023language}                                                                               \\ 
LLAMA        & Large language model meta AI \citep{touvron2023llama}                   & MLLM        & Multimodal large language model                                                                                                                            \\ 
GLaM         & Generalist language Model \citep{du2022glam}                                   & DRL         & Deep reinforcement learning                                                                                                                                \\ 
GPT-2        & Generative pretrained transformer 2 \citep{radford2019language}                & DQN         & Deep Q-network \citep{mnih2015human}                                                                                                      \\ 
GPT-3        & Generative pretrained transformer 3 \citep{brown2020language}                  & PPO         & Proximal policy optimization \citep{schulman2017proximal}                                                                      \\ 
GPT-4        & Generative pretrained transformer 4 \citep{openai2023gpt4}                     & Gato        & A aeneralist agent \citep{reed2022generalist}                                                                                             \\ 
Chat-GPT     & Chat generative pretrained transformer \citep{chatgpt}                         & AdA         & Adaptive agent \citep{team2023human}                                                                                                      \\ \hline
\end{tabular}
}
\end{table*}

To mitigate the aforementioned issues, many approaches have been proposed, including transfer learning \citep{weiss2016survey, zhuang2020comprehensive}, few-shot learning \citep{wang2020generalizing, yang2022survey}, label-efficient learning \citep{zhou2018brief,li2023label}, among others. Specifically, transfer learning has emerged as a widely adopted approach in agricultural applications and beyond \citep{espejo2020towards, chen2022performance, abbas2021tomato, dang2022deepcottonweeds}, which leverages DL models pre-trained on large-scale image datasets, such as ImageNet \citep{deng2009imagenet}, Microsoft COCO \citep{lin2014microsoft}, and PlantCLEF2022 \citep{goeau2022overview}. The pre-trained models are then fine-tuned for specific tasks. For instance, 35 state-of-the-art DL models are evaluated for weed classification using transfer learning approach \citep{chen2022performance}. Few-shot learning, on the other hand, focuses on training models to quickly adopt new tasks or domains using only a limited number of labeled examples, by leveraging prior knowledge or meta-learning approaches. In contrast, label-efficient learning \citep{li2023label} employs weak supervision or no supervision techniques to mitigate the effects of laborious and time-consuming labeling challenges. However, these methods are often pre-trained on data from a single domain and modality, which restricts their adaptability to other domains and applications \citep{raffel2020exploring}. The above observations highlight the importance of incorporating diverse data sources to enhance their generalization capabilities across different domains and applications. 

Large pre-trained models are also referred to as FMs \citep{bommasani2021opportunities, moor2023foundation, mai2023opportunities} designed to produce a wide and general variety of outputs, undergo extensive training on vast and diverse datasets. FMs are enabled to tackle numerous downstream tasks, such as speech recognition and natural language understanding \citep{radford2019language, brown2020language, chatgpt, ouyang2022training, touvron2023llama}, computer vision \citep{rombach2022high, ramesh2022hierarchical, kirillov2023segment, kang2023scaling}, and decision making \citep{reed2022generalist, team2023human}. FMs are generally trained on large-scale datasets from diverse domains and modalities through self-supervised learning \citep{liu2021self}, and are capable of handling a wide range of applications from various domains with minimum fine-tuning and no/little task-specific labeled data \citep{bommasani2021opportunities, moor2023foundation, mai2023opportunities}. For example, the Generative Pre-trained Transformer 3 language model (GPT-3, \citep{brown2020language}) demonstrates a breakthrough in in-context learning, enabling the model to perform novel tasks without explicit training, by leveraging text explanations (known as ``prompts'') containing just a few examples. Subsequently, ChatGPT \citep{chatgpt}, a derivative of  GPT-3, is developed and publicized by OpenAI for chat-based interactions and has brought about a transformative evolution in natural language processing, enabling unprecedentedly engaging conversational experiences. In the computer vision domain, FMs such as the segment anything model (SAM, \citep{kirillov2023segment}), trained on over 1 billion masks on 11M licensed and privacy-respecting images, can solve a range of downstream segmentation problems with new image distributions and tasks with zero-shot generalization where a model is trained to recognize classes or categories that it has never seen during training \citep{yang2022survey}. DeepMind Ada \citep{team2023human}, equipped with a customized Transformer architecture \citep{vaswani2017attention} using distillation with a teacher-student approach \citep{schmitt2018kickstarting}, brings FMs to reinforcement learning (RL) that generalizes to new tasks in a few-shot setting. Despite the aforementioned advances, FMs' adoption in agriculture has received scarce attention.

As such, this study aims to explore the potential of developing and applying FMs for agricultural applications. The main contributions and the technical advancements of this paper are summarized as follows.
\begin{itemize}
    \item Recent FMs (Section~\ref{sec:fms}) in the general computer science (CS) domain are systematically reviewed and categorized into four key categories: language FMs, vision FMs, multimodal FMs, and reinforcement learning FMs.
    \item The process of developing AFMs (Figure~\ref{fig:pipeline}) is outlined and their potential applications (Section~\ref{sec:afms}) in smart agriculture are discussed, shedding light on their diverse capabilities.
    \item This study also identifies and discusses the unique challenges and potential pitfalls associated with developing AFMs (Section~\ref{sec:dis}), including model training, validation, and deployment.
\end{itemize}

This comprehensive examination of AFMs can serve as a valuable resource for newcomers to the field, as well as experienced researchers seeking to innovate within the agricultural space.

\section{Foundation Models (FMs)}
\label{sec:fms}
In this section, both the strengths and weaknesses of FMs are first covered, giving readers a balanced view of the topic. Then, the different kinds of FMs, their distinct characteristics, and their applicability to various domains or tasks are explored.

\subsection{Pros and Cons of FMs}
FMs represent a significant paradigm shift in the field of AI. In contrast to traditional AI models that are trained on domain-specific data for specialized tasks, FMs are pre-trained on a vast scale of data, spanning multiple domains and modalities \citep{bommasani2021opportunities, moor2023foundation, mai2023opportunities}. The training process allows these models to learn a broad base of knowledge and skills, much like the foundation of a building that supports all the structures above it. Once trained, these FMs can accomplish versatile tasks with just a little fine-tuning. This is analogous to the way humans transfer knowledge and skills from one area to another \citep{moor2023foundation}. They have led to breakthroughs in a variety of fields, from natural language processing to computer vision and beyond. The pros of FMs are summarized as follows:
\begin{enumerate}
    \item \textbf{Pre-trained Knowledge}: By training on vast and diverse datasets, FMs possess a form of ``general intelligence'' that encompasses knowledge of the world, language, vision, and their specific training domains.
    \item \textbf{Fine-tuning Flexibility}: FMs demonstrate superior performance to be fine-tuned for particular tasks or datasets, saving the computational and temporal investments required to train extensive models from scratch.
    \item \textbf{Data Efficiency}: FMs harness their foundational knowledge, exhibiting remarkable performance even in the face of limited task-specific data, which is effective for scenarios with data scarcity issues. 
\end{enumerate}

Given the distinct advantages of FMs over traditional AI models, they exhibit substantial promise for agricultural applications. Some compelling examples of applying LLMs and FMs in the agricultural domain can be found in Section~\ref{sec:afms}.  However, despite their progress, FMs are not without their unique set of challenges:
\begin{enumerate}
    \item \textbf{Computational Cost}: The foundational training of these models demands considerable computational resources, often beyond the reach of individuals or firms with constrained budgets.
     \item \textbf{Generalization v.s. Specialization}: Although FMs offer commendable generalization capabilities across diverse tasks, there exist scenarios where task-specific models deliver superior performance.
     \item \textbf{Deployment Difficulties}: The considerable size of many FMs poses challenges for deployment on edge devices or other systems with restricted computational and storage capabilities.
    \item \textbf{Transparency and Interpretability}: The complexity and scale of FMs can hinder clear understanding and interpretation, making them opaque in certain applications.
    \item \textbf{Ethical Concerns}: Given their expansive training datasets, there's potential for these models to inadvertently perpetuate and magnify existing societal biases.
\end{enumerate}

For an in-depth discussion on deploying FMs in agricultural contexts, please refer to Section~\ref{sec:dis}.

\subsection{Categories of FMs}
\begin{figure}[!ht]
    \centering
{\includegraphics[width=.5\linewidth]{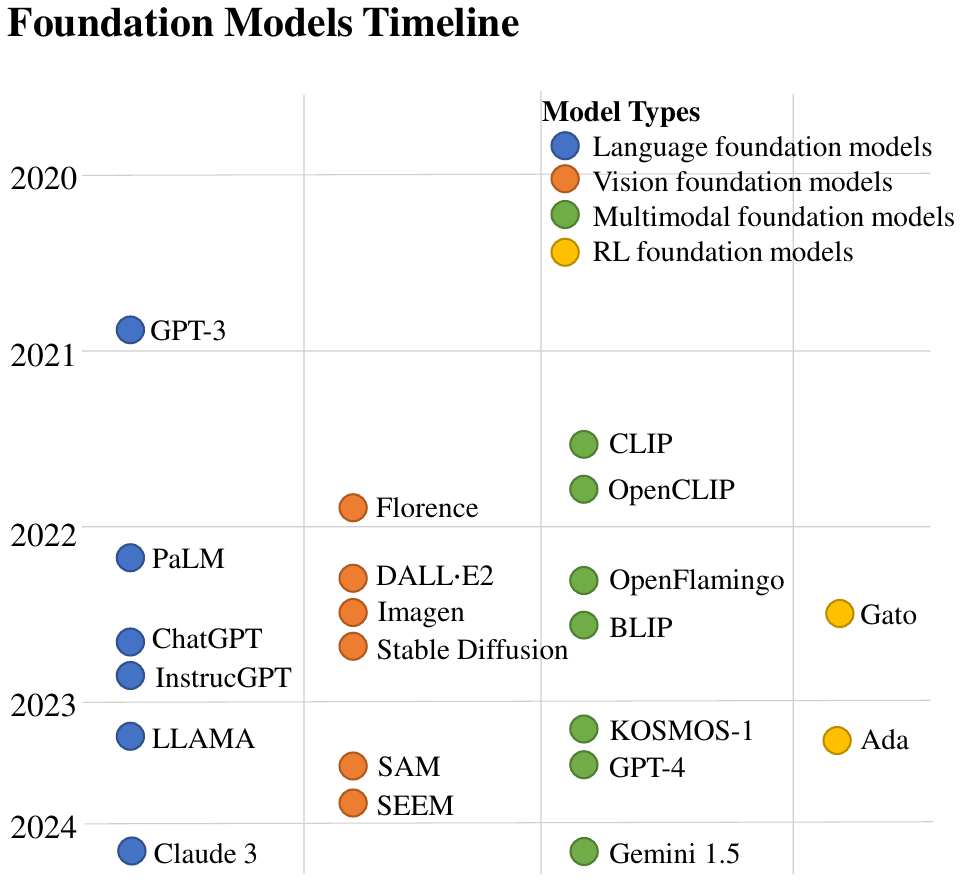} }
    \caption{Timeline of FMs.}
    \label{fig:fms}
    \vspace{-10pt}
\end{figure}

In this subsection, a comprehensive review of LLMs \& FMs within the computer science (CS) domain is presented, and the models are categorized into four distinct types based on the primary data modality and learning paradigm they address: language FMs (LFMs), vision FMs (VFMs), multimodal FMs (MFMs), and reinforcement learning FMs (RLFMs) (see Figure~\ref{fig:fms}). Readers who are familiar with these concepts or more interested in agricultural applications may skip this section and refer to Figure~\ref{fig:fms} for a concise summary.

\subsubsection{Language foundation models (LFMs)} \label{sec:lfms}
Over the past few years, the field of computational natural language has seen a complete transformation \citep{devlin2018bert, brown2020language} due to the rise of LLMs. Essentially, language modeling \citep{nadkarni2011natural} involves predicting the subsequent element in a given sequence of tokens, relying on a self-supervised objective that requires no manual labeling beyond a raw text corpus. 

The \underline{B}idirectional \underline{E}ncoder \underline{R}epresentations from  \underline{T}ransformers (BERT, \citep{devlin2018bert}) is one of the most prominent models, with two versions: $\text{BERT}_\text{BASE}$ and $\text{BERT}_\text{LARGE}$, containing 110M and 340M parameters respectively. By simply incorporating an additional output layer for fine-tuning, the pre-trained BERT model can be adapted to generate cutting-edge models for a variety of tasks, such as question answering and language inference. Remarkably, this can be achieved without requiring substantial modifications to the task-specific architecture, even surpassing the performance of many dedicated task-specific architectures. Subsequently, Google proposes the \underline{P}athways \underline{L}anguage \underline{M}odel (PaLM, \citep{wei2022chain}) and Large Language Model Meta AI (LLAMA, \citep{touvron2023llama}), further advancing the field of natural language processing. PaLM, with 540B parameters, is designed with a dense decoder-only Transformer model trained with the Pathways system \citep{barham2022pathways} to enable efficient training. PaLM surpasses the few-shot performance of prior large models, such as GLaM \citep{du2022glam}, GPT-3 \citep{brown2020language}, and LLAMA \citep{touvron2023llama}, on numerous very difficult tasks, such as nature language inference \citep{bowman2015large}, in-context reading comprehension \citep{jenkins2003sources}, and question answering \citep{kwiatkowski2019natural}. 

Participating in the development of large language models, OpenAI has proposed noteworthy contributions such as \underline{G}enerative \underline{P}retrained \underline{T}ransformer 2 (GPT-2, \citep{radford2019language}), 3 (GPT-3, \citep{brown2020language}), and 4 (GPT-4, \citep{openai2023gpt4}), having grown progressively more powerful, with 1.5 billion, 175 billion, and a staggering number of parameters, respectively, and setting remarkable benchmarks in language processing. ChatGPT \citep{chatgpt}, a specialized version of these models, has been extensively fine-tuned for human-like conversation. By adjusting the final output layer, the pre-trained GPT models can be tailored to deliver exceptional performance across a wide range of tasks, such as text generation, translation, and dialogue systems, outperforming many task-specific models without requiring extensive modifications to their foundational architecture. Recently, OpenAI has shown that InstructGPT \citep{ouyang2022training} with only 1.3B parameters are much better at following users' instructions than GPT-3, by using an existing technique called reinforcement learning from human feedback (RLHF, \citep{christiano2017deep}). More recently,  the development of Claude 3\footnote{Claude 3 website: \url{https://www.anthropic.com/news/claude-3-family}.} represents the cutting-edge in AI systems, encompassing undergraduate-level expert knowledge (MMLU), graduate-level expert reasoning (GPQA), basic mathematics (GSM8K), and beyond. It demonstrates nearly human-like levels of comprehension and fluency across intricate tasks, marking the forefront of general intelligence.

With the continuous growth in data availability and computing power, language modeling has become a trustworthy method for crafting ever more powerful models.

\subsubsection{Vision foundation models (VFMs)} \label{sec:vfms}
Computer vision, a field dedicated to understanding and analyzing visual data, has traditionally relied on task-specific models tailored to specific objectives. These models, such as ResNet \citep{he2016deep} for image classification, Faster RCNN \citep{ren2015faster} for object detection, Detectron \citep{Detectron2018} for instance segmentation, and DCGAN \citep{radford2015unsupervised} for image generation, have made significant contributions to advancing visual perception tasks. However, it is important to note that these task-specific models are designed to excel in their respective applications and may not be directly applicable to other computer vision tasks \citep{bommasani2021opportunities}.

With the emergence of vision FMs, there is a paradigm shift towards more versatile and general-purpose models that can tackle multiple vision tasks simultaneously. These FMs leverage deep learning techniques, and diverse, and large-scale training data to offer a unified framework for various computer vision tasks, enabling researchers to explore new frontiers in visual understanding and analysis \citep{bommasani2021opportunities}. For example, Florence \citep{yuan2021florence}, a newly introduced computer vision FM, offers an extensive range of capabilities for visual understanding. By leveraging universal visual-language representations from vast image-text datasets, Florence seamlessly adapts to various computer vision tasks including classification, retrieval, object detection, visual question answering (VQA), image captioning, video retrieval, and action recognition. Notably, Florence achieves outstanding performance in transfer learning scenarios, excelling in fine-tuning, linear probing, few-shot transfer, and zero-shot transfer for novel images and objects. Recently, the \underline{S}egment \underline{A}nything \underline{M}odel (SAM, \citep{kirillov2023segment}) is introduced by Meta AI for image segmentation. With its flexible and efficient architecture, SAM achieved impressive zero-shot performance on numerous segmentation tasks \citep{zhang2023asurvey, ma2023segment, yang2023sam, tang2023can}. To promote research in FMs for computer vision, the SA-1B dataset, the largest of its kind, providing over 1 billion masks from 11 million images, is open-sourced\footnote{SAM project \url{https://segment-anything.com/}}. Inspired by SAM, the Segment Any Anomaly + (SAA+, \citep{cao2023segment}) is proposed for zero-shot anomaly segmentation with hybrid prompt regularization to improve the adaptability of modern FMs. Furthermore, in \cite{tang2023can}, the authors show that SAM shows promising performance in camouflaged object detection (COD) tasks, in which objects are seamlessly integrated into their surroundings. In \cite{yang2023sam}, SAM demonstrates superior performance in the poultry industry, specifically in the context of segmenting cage-free hens, outperforming SegFormer \citep{xie2021segformer} and SETR \citep{liu2022setr} in both whole and part-based chicken segmentation. The \underline{S}egmenting \underline{E}verything \underline{E}verywhere \underline{M}odel (SEEM, \citep{zou2023segment}) represents a significant advancement in the field of image segmentation. It sets a new benchmark for simultaneously segmenting everything everywhere in an image, including semantic, instance, and panoptic segmentation \citep{kirillov2019panoptic}. Notably, SEEM supports a wide range of prompts, including visual, textual, and referring region prompts, which can be combined in various ways to enable versatile and interactive referring segmentation. This model represents a breakthrough that pushes the boundaries of what is achievable in comprehensive image segmentation tasks.

Vision FMs have also demonstrated outstanding performance in the field of image generation, showcasing their ability to produce high-quality, diverse, and realistic images that push the boundaries of computer-generated visuals. The current attempts in this area can be categorized into two main branches: generative adversarial networks (GANs, \citep{goodfellow2020generative}) and diffusion probability models (diffusion models (DMs \citep{yang2022diffusion}). Stable Diffusion \citep{rombach2022high}, a latent text-to-image DM, is introduced to generate detailed images conditioned on text descriptions. The model leverages a frozen CLIP ViT-L/14 text encoder \citep{radford2021learning} to incorporate text prompts, combining an 860M UNet and a 123M text encoder. The Stable Diffusion model achieves state-of-the-art scores in various tasks, such as image inpainting and class-conditional image synthesis. Around the same time, OpenAI released DALL$\cdot$E2 \citep{ramesh2022hierarchical}, another model for generating realistic images based on text inputs. DALL$\cdot$E2 consists of a prior stage, responsible for generating a CLIP \citep{radford2021learning} image embedding from a given text caption, and a decoder stage, which generates an image conditioned on the generated image embedding. This architecture enables language-guided image manipulations in a zero-shot manner. Moreover, \cite{saharia2022photorealistic} introduces Imagen, which exhibits superior sample quality and image-text alignment compared to previous works such as VQ-GAN+CLIP \citep{crowson2022vqgan}, Stable Diffusion \citep{rombach2022high}, GLIDE \citep{nichol2021glide} and DALL$\cdot$E2 \citep{ramesh2022hierarchical} with an efficient U-Net diffusion network architecture and powerful large transformer language models. 

While the previous works discussed above have primarily focused on diffusion models (DMs), GigaGAN \citep{kang2023scaling} represents a significant advancement in the realm of image generation by adopting a generative adversarial network (GAN) approach. GigaGAN offers significantly faster inference time compared to DM-based approaches \citep{rombach2022high, ramesh2022hierarchical, saharia2022photorealistic} while supporting various applications like latent interpolation, style mixing, and vector arithmetic operations. Building upon the StyleGAN2 framework \citep{karras2020analyzing}, GigaGAN enhances training stability and model capacity by retaining a bank of filters and employing a combination of self-attention (image-only) and cross-attention (image-text) with the convolutional layers for improved performance. Additionally, the authors introduce multi-scale training and multi-stage approaches to enhance image-text alignment and generate low-frequency details. GigaGAN, with 1.0B parameters, achieves a zero-shot Fréchet Inception Distance (FID) of 9.09 on the COCO2014 dataset, outperforming Imagen (3.0B) and DALL$\cdot$E2 (5.5B) in terms of image quality evaluation.

Looking ahead, VFMs are expected to continue evolving and pushing the boundaries of what is possible in computer vision. As research progresses, it can be anticipated that such models will be further advanced with improved capabilities in terms of image understanding, scene understanding, and content generation. Furthermore, the fusion of vision models with other modalities, such as language and audio, is likely to lead to the emergence of more powerful and versatile multimodal models (Section~\ref{sec:mfm}).

\subsubsection{Multimodal foundation models (MFMs)} \label{sec:mfm}
MFMs represent a new class of AI models that leverage and integrate multiple types of data inputs, like text, video, audio, and images, which is a distinct departure from their non-multimodal counterparts that are constrained to processing a singular data type. By capitalizing on the richness and diversity of these data types, MFMs can attain a more holistic understanding of the context. This, in turn, empowers them to execute more nuanced and precise tasks, such as translating text into visual representations, identifying objects in images based on textual prompts, and creating image captions. As a result, these models can fuel a vast array of real-world applications \citep{bommasani2021opportunities, moor2023foundation, mai2023opportunities}. 

One notable example is OpenAI's CLIP (\underline{C}ontrastive \underline{L}anguage-\underline{I}mage \underline{P}retraining, \citep{radford2021learning}), which has garnered significant attention for its ability to learn joint representations of images and text. It utilizes a strategy known as contrastive learning \citep{zhao2021contrastive, jaiswal2020survey}, which entails learning to associate accurate text-image pairs while dissociating inaccurate ones. Consequently, CLIP excels at tasks such as zero-shot learning, wherein it can classify new images based on textual descriptions and vice versa. Its multimodal nature renders it highly advantageous in diverse domains, such as computer vision, natural language processing, and digital content creation. Building on this foundation, OpenCLIP \citep{Ilharco_Open_Clip_2021}, a model trained on up to 2B image-text pairs, is proposed to investigate scaling laws \citep{kaplan2020scaling} for CLIP \citep{radford2021learning} model with public datasets. OpenCLIP identifies power scaling laws through various downstream tasks, such as zero-shot retrieval, classification, linear probing, and end-to-end fine-tuning, concluding that the training distribution plays a key role in scaling laws. The new Gemini 1.5\footnote{Gemini 1.5 website: \url{https://deepmind.google/technologies/gemini/\#introduction}.} developed by Google delivers dramatically enhanced performance with a more efficient architecture compared to its previous version \citep{team2023gemini}.

However, a majority of FMs are typically engineered for either understanding-based or generation-based tasks. To address this issue, BLIP \citep{li2022blip}, a new Vision-Language Pre-training (VLP) framework, is introduced for both vision-language understanding and generation tasks. To effectively utilize the noisy web data by bootstrapping the captions, a captioner is employed to generate synthetic captions, subsequently, a filter is adopted to remove the noisy ones. BLIP achieves state-of-the-art results across various vision-language tasks, such as image-text retrieval, image captioning, and VQA. Similarly, OpenFlamingo \citep{Alayrac2022FlamingoAV} is trained on large-scale multimodal web corpora with arbitrarily interleaved text and images for a variety of image and video tasks via few-shot learning, such as image captioning and multiple-choice VQA \citep{Alayrac2022FlamingoAV}. KOSMOS-1 \citep{huang2023language}, a Multimodal Large Language Model (MLLM), is subsequently introduced to process multimodal inputs, adhere to instructions, and perform in-context learning for language tasks and multimodal tasks, such as multimodal dialogue, image captioning, VQA, and image recognition with descriptions. Impressively, KOSMOS-1 demonstrates exemplary performance across various settings, such as zero-shot, few-shot, and multimodal chain-of-thought prompting, on various tasks without necessitating any gradient updates or fine-tuning. Built on previous versions, GPT-4 \citep{openai2023gpt4}, a large-scale and multimodal model is introduced by OpenAI to accept both image and text inputs and produce text outputs. 

The future of MFMs is set to be exciting, with further advancements anticipated in several areas. They will likely include more sophisticated integration of various data modalities and enhanced transfer learning for broader applicability across different domains \citep{bommasani2021opportunities}. Furthermore, efforts are expected to be directed towards minimizing data bias from different domains, improving fairness, and developing energy-efficient AI models \citep{bommasani2021opportunities}. 

\subsubsection{Reinforcement learning foundation models (RLFMs)}\label{sec:rlfms}
Over the past decade, the field of decision-making has undergone a significant revolution due to the emergence of RL algorithms integrated with advanced DL/ML algorithms. These models, e.g., deep RL (DRL) algorithms like DQN \citep{mnih2015human} and PPO \citep{schulman2017proximal}, have enabled RL agents to learn complex behaviors and achieve remarkable performance in various domains, such as game playing \citep{mnih2013playing, mnih2015human, vinyals2017starcraft}, robot control \citep{kober2013reinforcement, lee2020learning}, and agricultural applications \citep{bu2019smart, binas2019reinforcement, gandhi2022deep, gautron2022reinforcement}. By interacting with the environment, these models learn optimal decision-making through trial and error. However, due to their reliance on learning tasks from scratch without comprehensive knowledge of vision, language, or other datasets, these methods often face challenges in terms of generalization and sample efficiency \citep{yang2023foundation}. 

There are increasing attempts to combine FMs and sequential decision-making to tackle complex real-world problems with better generalization. Gato \citep{reed2022generalist}, a multi-modal, multi-task, and multi-embodiment generalist RL policy, is proposed by Google DeepMind with a 1.2B parameter decoder-only transformer architecture, trained on 604 distinct tasks with varying modalities, observations, and action specifications. Gato is able to play Atari, caption images, chat, and stack blocks with a real robot arm, using the same network and weights with little fine-tuning. Subsequently, AdA \citep{team2023human} proposes by Google DeepMind with 265M parameters that learn new tasks in an extended and 3D environment simulation as fast as humans based on a modified transformer architecture to store more information and enable efficient training. 

Leveraging world knowledge from FMs, sequential decision making can solve tasks faster and generalize better \citep{yang2023foundation}.  FMs for decision making are still in their infancy and still facing significant challenges including the discrepancy in data modalities, uncertainties surrounding environment and task structures, and the absence of certain components in existing  FMs and decision-making paradigms. Further advancements are necessary to address these challenges and unlock the full potential of  FMs in decision making \citep{yang2023foundation, reed2022generalist}.

\section{Agricultural Foundation Models (AFMs)}
\label{sec:afms}
In this section, the potential for creating FMs in agriculture is first examined, i.e., AFMs. Then current applications of these models in the realm of smart agriculture are presented, followed by a discussion on emerging areas where these FMs are most applicable.

\subsection{Developing FMs for Agriculture}
Figure~\ref{fig:pipeline} presents a comprehensive overview of the pipeline for developing (multimodal) FMs in agriculture. This process principally encompasses stages such as data collection, dataset curation, model training, and downstream fine-tuning.

The development of MFMs for agricultural applications commences with the imperative phase of data collection and curation \citep{bommasani2021opportunities}. This process necessitates the assembly of heterogeneous data spanning a range of modalities, including visual (e.g., images and videos indicating the status of crops, pests, and disease symptoms), textual (e.g., human instructions and pest identification documents), meteorological (e.g., temperature, precipitation, and wind speed measurements), and possibly auditory data (e.g., distinctive insect or animal sounds). Given the magnitude of the data involved, meticulous annotation and curation are indispensable to discriminate meaningful patterns and correlations. Such data labeling could encompass tasks such as the classification of crop and weed species, the annotation of disease symptoms present on plant foliage, or the categorization of weather conditions.

Subsequent to data curation is the model training phase. Capitalizing on contemporary advancements in ML and AI, such as transformer architectures \citep{vaswani2017attention} and convolutional neural networks \citep{gu2018recent}, the model is trained to process and interpret multimodal inputs. These AI models are designed to associate visual depictions with their corresponding textual labels, comprehend textual descriptions, and generate an inclusive understanding of the agricultural ecosystem. Notably, the training of these models may be computationally demanding, which necessitates the use of advanced hardware and optimization algorithms to enhance performance, which will be discussed in Section ~\ref{sec:training_fms}.

The culmination of this process is the implementation of these trained models in downstream agricultural applications, such as plant health monitoring \citep{xu2022style}, crop and weed management \citep{hu2021deep,dang2023yoloweeds, rai2023applications}, fruit picking \citep{chu2021deep}, and precision livestock farming \citep{yang2023sam}. For instance, in relation to plant health, the models can be utilized to identify early indicators of disease or pest infestations, thus facilitating prompt interventions to reduce crop loss. For weed management, the models can assist in distinguishing weeds from crops, enabling precision farming equipment to selectively eradicate unwanted flora. Furthermore, in the domain of precision livestock farming, these models can be harnessed to monitor livestock health and behavior, optimize feeding schedules, and predict potential diseases. In the context of downstream tasks, transfer learning or fine-tuning \citep{weiss2016survey, zhuang2020comprehensive} of the FMs becomes paramount to adapt the generalized learning to specific agricultural applications, thereby enhancing model performance and efficiency. Through these varied applications, MFMs hold the potential to significantly transform the agricultural industry, thereby enhancing productivity, promoting sustainability, and ultimately boosting profitability.

\subsection{Agricultural applications}
In this subsection, the existing applications of integrating FMs discussed are explored in Section~\ref{sec:fms} into the field of smart agriculture.


\subsubsection{LFMs in agriculture}
LFMs, also referred to as LLMs and detailed in Section ~\ref{sec:lfms}, have transformed the realms of design, development, and deployment by enabling human-like interactions. Researchers are now increasingly integrating these LFMs into various stages of design and development for agricultural applications, as highlighted in \citep{stella2023can, lu2023agi}. Specifically, in \cite{stella2023can}, the authors discuss the process of integrating LLMs into the design phase of robotic systems, as depicted in Figure 2 from \cite{stella2023can}. They specifically illustrate the procedure to design a robotic gripper optimized for tomato picking. During the initial ideation phase, researchers consult LLMs, such as ChatGPT \citep{chatgpt}, to understand potential challenges and opportunities within the field. Based on this information, they select the most promising and intriguing pathways, subsequently narrowing down the design possibilities through further dialogue with the LLM. This collaborative interaction covers a wide spectrum of knowledge domains and various levels of abstraction, encompassing everything from conceptual thinking to technical implementation. Throughout this process, the human collaborator leverages the AI's knowledge to access information beyond their personal expertise. During the subsequent phase of the design process, which is more technically focused, these general directions need to be actualized into a tangible, fully functional robot. As of now, LLMs are not capable of producing complete computer-aided design
 (CAD) models, assessing code, or autonomously fabricating a robot. However, recent progress in AI research has indicated that these algorithms can provide substantial assistance in executing software \citep{chen2021evaluating}, mathematical reasoning \citep{wolfram2023wolfram}, and even in the generation of shapes \citep{ramesh2022hierarchical}. Although future AI methodologies will be capable of handling these tasks, at the moment, the technical implementation remains a team effort between AI models and humans. Finally, the human would fine-tune the code suggested by the LLM, finalize the CAD design, and oversee the actual fabrication of the robot. Figure 2(b) in  \cite{stella2023can} shows the primary outputs generated by the LLM, as well as the real-world application of the AI-designed robotic gripper used for crop harvesting.

The authors in \cite{lu2023agi} have deliberated on the prospective use cases of LLMs in organizing unstructured metadata, converting metadata from one format to another, and identifying potential errors during the data collection phase. They also envision the future generation of LLMs as exceptionally powerful tools for data visualization \citep{bubeck2023sparks}, offering invaluable assistance to researchers in extracting meaningful insights from vast volumes of phenotypic data. 
Furthermore, in \cite{tzachor2023large}, the authors utilize GPT to revolutionize agricultural extension services by simplifying scientific knowledge and delivering personalized, location-specific, and data-driven agricultural recommendations. In \cite{shutske2023harnessing}, Generative AI and LLMs, particularly ChatGPT, are investigated to elevate innovation and enhance agricultural safety, occupational health, and health promotion.

Moreover, researchers have leveraged LLMs for various plant science applications. For example, in \cite{yang2024pllama}, PLLaMa, an open-source LLM based on LLaMa-2, is developed specifically for plant science-related topics. Additionally, in \cite{kuska4685971ai}, an AI-Chatbots for Agriculture is created to aid in interpreting decision support models in plant disease management.

\subsubsection{VFMs in agriculture}

\begin{figure*}[!t]
\centering
\includegraphics[width=.80\linewidth]{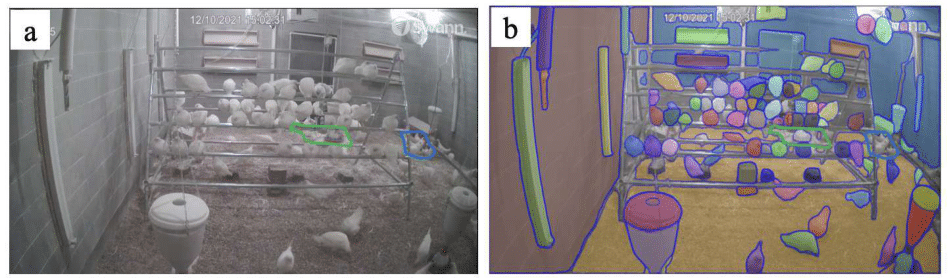}
\includegraphics[width=.80\linewidth]{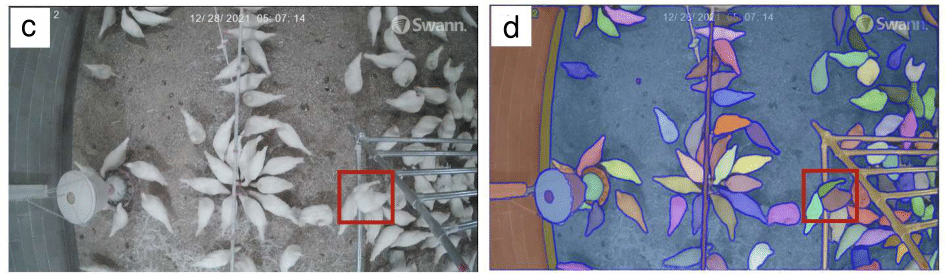}
\caption{Side-veiw (a,b) and top-view (c,d) views of chicken segmentation using SAM model in a research cage-free house proposed in \cite{yang2023sam}.}
\label{fig:hicken_segmentation}
\vspace{-10pt}
\end{figure*}

VFMs (as detailed in Section ~\ref{sec:vfms}), encompassing algorithms for core computer vision tasks like classification, detection, segmentation, and generative modeling, have revolutionized traditional computer vision methodologies. Trained on extensive image datasets, these VFMs can be efficiently adapted to novel domains, such as agriculture, with minimal fine-tuning and a limited set of labeled images. Recently, the agricultural community has shown a growing interest in VFMs, as seen in recent studies \citep{yang2023sam, williams2023leaf}. 

In \cite{yang2023sam}, SAM (Section ~\ref{sec:vfms}) is employed for chicken segmentation tasks in a zero-shot manner using both part-based segmentation and the use of infrared thermal images (Figure~\ref{fig:hicken_segmentation} ). Experimental results show that SAM outperformed the other VFMs algorithms, such as SegFormer and SETR, in both whole and partial chicken segmentation. For example, SAM achieves a mean Intersection over Union (mIoU) of 94.80\% in the chicken segmentation tasks, while SegFormer and SETR only achieve mIoUs of 43.22\% and 42.90\%, which greatly shows the effectiveness of SAM on the poultry segmentation tasks. In addition, a novel single-bird tracking algorithm, integrating SAM, YOLOX \citep{ge2021yolox} and ByteTracker \citep{zhang2022bytetrack}, is proposed for tracking the activities for individual birds across video frames by extracting the location information (i.e., bounding boxes provided by SAM) in each image and then tracking the birds over time with YOLOX and ByteTracker. Despite its promising performance, the tracking algorithm lacked end-to-end capability. The object tracking FMs \citep{yang2023track} show impressive performance in video object tracking and segmentation, requiring very little human participation, thereby holding the potential to serve as an end-to-end method.

In \cite{williams2023leaf}, an automatic leaf segmentation pipeline, named ``Leaf Only SAM'', is introduced for zero-shot segmentation of potato leaves. When compared with a fine-tuned Mask R-CNN \citep{he2017mask} model on the annotated potato leaf dataset, Leaf Only SAM demonstrates an average recall and precision of 63.2\% and 60.3\%, respectively. These results are lower than the Mask R-CNN model's average recall and precision of 78.7\% and 74.7\%, respectively. However, it's important to note that Leaf Only SAM does not necessitate any fine-tuning or an annotated dataset, thereby showing notable zero-shot generalization capabilities. These findings suggest that the direct application of FMs in a zero-shot manner may not always result in high performance due to potential distribution shifts \citep{chawla2021quantifying}. This will be further discussed in Section ~\ref{sec:df}.

\subsubsection{MFMs in agriculture}
MFMs (Section ~\ref{sec:mfm}) introduce a novel class of AI models. These leverage and incorporate various types of data inputs including text, video, audio, and images. The agricultural community has begun to delve into multimodal learning within agricultural applications \citep{bender2020high, garg2021towards, parr2021multimodal, cao2023cucumber, dhakshayani2023m2f} to enhance performance through the utilization of multimodal data. For instance, \cite{bender2020high} open-sources a multimodal dataset designed for agricultural robotics, which is gathered from cauliflower and broccoli fields, to facilitate robotics and machine learning research activity in agriculture. The dataset includes a variety of data such as stereo color, thermal and hyperspectral imagery, and crucial environmental details like weather and soil conditions. In \cite{cao2023cucumber}, a multi-modal language model based on image-text-label information is proposed for cucumber disease recognition. They effectively merge label information across multiple domains by employing image-text multimodal contrastive learning and image self-supervised contrastive learning.  This methodology helps measure the distance of samples within the common image-text-label space. The experimental results reveal that this novel approach attains a recognition accuracy rate of 94.84\% on a modestly sized multimodal cucumber disease dataset.

MFMs have also been discussed in the agriculture domain. \cite{lu2023agi} investigates the potential of GPT-4, a multimodal language and vision model (as referenced in ChatGPT, Section~\ref{sec:lfms}), in achieving AGI. In \cite{stella2023can}, LLMs (e.g., ChatGPT) are utilized in the design phase of robotic systems for tomato picking. 
In \cite{tan2023promises}, GPT-4 is applied for geographical, environmental, agricultural, and urban planning applications, such as crop type identification.

However, it is noteworthy that current applications are predominantly based on text-image data, and their use is often confined to question-answering tasks. Indeed, there is a conspicuous gap in applications related to agricultural robotics where inputs such as images, text, or voice (e.g., human instructions), and depth information (e.g., from LiDAR or laser sensors) are used. These robots, which are typically designed for tasks like fruit harvesting or crop monitoring \citep{tao2017automatic}, could greatly benefit from the application of MFMs, which will be further discussed in Section ~\ref{sec:ar}.

\subsubsection{RLFMs in agriculture}
DRL \citep{arulkumaran2017deep, henderson2018deep} is an emerging field that fuses deep learning with reinforcement learning to tackle decision-making tasks. Its impressive performance has been demonstrated across a broad spectrum of applications, from computer games \citep{mnih2013playing, mnih2015human}, to intelligent transportation systems \citep{chu2019multi, chen2023deep}, and robot control \citep{gu2017deep, lee2020learning}. Recently, the agricultural community has shown heightened interest in DRL \citep{binas2019reinforcement, gandhi2022deep}, leading to a host of innovative applications such as crop management \citep{gautron2022reinforcement, din2022deep}, agriculture robots \citep{yang2022intelligent}, and smart irrigation systems \citep{zhou2020intelligent}. However, due to their reliance on learning tasks from scratch without comprehensive knowledge of vision, language, or other datasets, these methods often face challenges in terms of generalization and sample efficiency \citep{yang2023foundation}. 

Although there currently exist no published papers specifically focusing on the application of RLFMs (Section~\ref{sec:rlfms}) in agricultural scenarios, the field nonetheless presents vast untapped opportunities for exploration and innovation. For instance, Gato \citep{reed2022generalist} and AdA \citep{team2023human}, discussed in Section ~\ref{sec:rlfms}, exhibit promising potential for handling decision tasks that involve multimodal, multi-task contexts, paving the way for innovative agriculture applications in the future.

Table~\ref{tab: app sum} summarizes the applications of LLMs and FMs in agriculture.

\begin{table*}[!ht]
\renewcommand{\arraystretch}{1.4}
\centering
\caption{Applications of LLMs and FMs in agriculture.}
\label{tab: app sum}
\resizebox{0.95 \textwidth}{!}{%
\begin{tabular}{c|c|c|c|c}
\hline
Model Type & Reference                                & Problem                   & Method                 & Applications                               \\ \hline
\multirow{3}{*}{LFMs}                                 & \cite{stella2023can}    & Robotic picking           & ChatGPT                & Agriculture robots   \\
& \cite{tzachor2023large}    & Agricultural recommendations           & GPT                & Agricultural extension \\
& \cite{lu2023agi}    & Data processing           & GPT                & Organizing unstructured metadata \\
& \cite{yang2024pllama}    & Plant science        & LLAMA                &  Plant science-related tasks \\ 
& \cite{shutske2023harnessing}    & Agricultural safety \& health         & ChatGPT                &  Agricultural safety and health questions \\
& \cite{kuska4685971ai}    & Ai-Chatbots         & ChatGPT                &  Plant disease management \\ \hline
\multirow{2}{*}{VFMs}                & \cite{yang2023sam}      & Segmentation and tracking & SAM & Chicken segmentation, bird tracking \\  
                                     & \cite{williams2023leaf} & Segmentation              & SAM & Automatic leaf segmentation               \\ \hline
\multirow{2}{*}{MFMs}                & \cite{cao2023cucumber}  & Disease recognition      & Image-text data       & Cucumber disease recognition              \\  
                                     & \cite{stella2023can}    & Robotic picking           & Image-text data        & Tomato picking                            \\ 
& \cite{tan2023promises}    & Remote Sensing        & GPT4      & Crop type
identification                           \\ \hline
RLFMs                                & -                                     &  -                         & -                       &         -                                  \\ \hline
\end{tabular}
}
\end{table*}

\subsection{Emerging opportunities}
In this subsection, some emerging areas in the agricultural space are discussed where these FMs are \textbf{most} applicable.

\subsubsection{Smart crop management}
To achieve sustainable farming, smart crop management \citep{eli2019applications, jaramillo2020sustainable} has evolved as a novel research direction. It integrates advanced methodologies from AI, IoT, and big data into crop management tasks, including plant growth monitoring \citep{lakshmi2017implementation}, disease detection \citep{xu2022style}, yield monitoring \citep{khanal2018integration, kayad2019monitoring}, and harvesting \citep{meshram2021machine}. Yet, developing such algorithms for precision crop management is complex and challenging, particularly when considering factors like diverse crop varieties, environmental dynamics, and limited or hard-to-acquire training datasets. Furthermore, these systems often necessitate diverse data inputs sourced from varying domains. For example, while a deep learning method is developed on satellite imagery to assess crop health, it simultaneously might need soil sensor data to understand underground moisture levels and market data to predict the optimal time for harvest based on pricing trends. Designing such algorithms requires deep research insights, making it challenging to smoothly incorporate knowledge from unrelated fields.

In light of the aforementioned challenges, FMs offer a promising solution. These models, trained on datasets spanning multiple domains, can be seamlessly adapted to new areas with minimal fine-tuning. They hold significant potential as tools for smart crop management. For instance, LFMs (Section ~\ref{sec:lfms}) empower researchers to tap into knowledge beyond their personal expertise, facilitating software development and design \citep{stella2023can}. Moreover, MFMs (Section ~\ref{sec:mfm}) integrate visual, textual, and sometimes auditory data, providing a comprehensive analytical lens, proving invaluable in deciphering intricate agricultural situations, notably in precision crop management.

\subsubsection{Smart plant breeding}
Smart plant breeding, a breakthrough approach, combines data from genotype, environment, and their interaction to optimize crop varieties \citep{xu2022smart, zhang2023smart}. Employing sophisticated ``omics'' technologies, AI, and big data, it constructs effective models to predict plant traits and performance (phenotype). The objective of this ``smart breeding'' is to promote genetic gains, shorten breeding cycles, and develop plants that are acclimatized to specific and evolving environments \citep{xu2022smart}. However, the accuracy in phenotype prediction is intrinsically tied to the processing of multidimensional or multimodal data, sourced from spatiotemporal omics, encompassing genotype, environment, and their interaction. Such a data-rich environment poses significant challenges to conventional machine learning methodologies \citep{xu2022smart}. 

In response to this complexity, MFMs (Section ~\ref{sec:mfm}) may emerge as a potential solution. These models, designed to handle and learn from varied data types across multiple domains, could significantly improve the accuracy and efficiency of predicting plant phenotypes, thereby pushing the field of smart plant breeding toward unprecedented possibilities.

\subsubsection{Smart livestock farming}
Smart livestock farming is an innovative, automated approach that fuses cutting-edge technologies such as the Internet of Things (IoT), deep learning, and technology for intelligent control of livestock production  \citep{islam2019smart, astill2020smart, zhang2021wearable, farooq2022survey}. This includes the automation of critical processes such as food and water supply, egg collection, livestock activity monitoring, and maintenance of precise environmental conditions. The primary objective of this innovative methodology is to enhance operational efficiency in livestock farms, leading to considerable savings in labor, maintenance, and material costs \citep{choukidar2017smart}. Smart livestock farming, while innovative and promising, also presents challenges. The development and implementation of such an intelligent system require a multidisciplinary understanding that spans the realms of advanced technologies, animal science, environmental science, and data analytics, among others \citep{islam2019smart}. This integrative approach involves handling and processing voluminous and varied data across multiple domains. Overcoming these challenges would require strategic collaborations, expertise across domains, and harnessing cutting-edge tools in AI and data analytics.

Given these complex challenges, FMs emerge as a potential solution to facilitate the realization of smart livestock farming \citep{yang2023sam}. These comprehensive models, built on large-scale multimodal data and incorporated knowledge from various domains, can effectively manage the complications of multi-dimensional information and bridge the gaps between different disciplines \citep{bommasani2021opportunities}. FMs, with their ability to learn and generalize from large datasets, could provide valuable insights, enhance decision-making, and enable more precise management, thereby contributing significantly to the advancement of intelligent and efficient livestock farming systems.

\subsubsection{Precision aquaculture farming}
Precision aquaculture farming \citep{o2019precision,fore2018precision,  antonucci2020precision}, also known as smart aquaculture or precision fish farming, is a modern approach to fish and shellfish farming that integrates advanced technologies like IoT and data-driven techniques to optimize production efficiency, environmental sustainability, and animal welfare. It is often characterized by hundreds of interconnected sensors that store and serve data to gather essential information such as temperature, oxygen, weather, chlorophyll, fish tags, imagery data, and 3D sensor data, which can be utilized for monitoring fish variables (species, health, etc.), behavioral analysis, feeding decisions, and water quality prediction \citep{yang2021deep}. Recently, DL has emerged as a promising solution to address the problems in precision aquaculture farming. For example, in \cite{maaloy2019spatio}, a dual-stream recurrent network was adopted to classify fish feeding status for salmon, yielding 80\% feeding classification accuracy. However, these DL approaches rely heavily on large-scale and labeled datasets for training \citep{yang2021deep}.  Underwater conditions can be adverse, with poor illumination and low visibility in turbid water, as well as cluttered backgrounds, making it difficult to acquire high-quality data. Besides, it is very challenging to integrate, handle, and process numerous and varied data across multiple domains. 

Fortunately, FMs have become an encouraging solution to facilitate the realization of precision aquaculture farming, representing the unique capability to handle diverse data from multiple domains through few-short or zero-short learning. For example, generation-based FMs can be harnessed for data creation, while RLFMs can be utilized for decision-making in fish farming. The application of FMs could significantly enhance the accuracy and efficiency of associated applications in precision aquaculture farming.

\subsubsection{Agricultural robots} \label{sec:ar}
Agricultural robots, often referred to as ``AgriRobots'', are at the forefront of transforming modern agriculture \citep{bechar2016agricultural}. AgriRobots can automate various farming tasks such as planting, weeding, harvesting, and crop health monitoring \citep{fountas2020agricultural}. For instance, AI-guided robots can identify and remove weeds with precision, significantly reducing the use of herbicides. In smart orchards, AI-guided robots equipped with a variety of sensors, including image sensors, LiDAR, and proximity sensors, autonomously pick ripe fruits, thereby optimizing harvesting efficiency and minimizing labor costs. Similarly, drones equipped with advanced imaging sensors can monitor fields for signs of disease or pests, enabling early intervention and reducing yield losses. Similar to the contexts of smart breeding and smart livestock farming, the development and implementation of AgriRobots require multidisciplinary knowledge including robotics, AI, plant science, environmental science, and data analytics. This mixed approach involves handling and processing a significant amount of data from various domains, making it a complex endeavor \citep{bechar2016agricultural, fountas2020agricultural}.

The above challenges can be guided effectively by the potential application of FMs. These models, trained on large-scale multimodal data, can provide the robust predictive capabilities, which are necessary for advanced farming operations. For instance, RLFM (Section ~\ref{sec:rlfms}) could enable precision control of AgriRobots \citep{reed2022generalist, team2023human}, while LFM (Section ~\ref{sec:lfms}) could provide valuable insights during the design process of AgriRobots \citep{stella2023can}. By incorporating insights from diverse domains, FMs can offer an integrated solution, driving the development of more effective and efficient agricultural robots, and thus, revolutionizing the future of agriculture.

However, constructing an AFM from scratch presents significant challenges. Firstly, initial data collection requires a large and diverse dataset, which requires laborious efforts and potentially high costs \citep{lu2020survey}. The subsequent data labeling and curation require meticulousness and expertise. Additionally, the process of training models demands substantial computational resources, thus presenting impediments for individuals or entities with constrained resources. Lastly, fine-tuning the model for specific agricultural applications can be challenging, requiring a balance between generalizability and specificity. A detailed discussion of these challenges will be presented in Section ~\ref{sec:dis}.

\section{Challenges and Outlook}
\label{sec:dis}
Although FMs have demonstrated impressive performance in various agriculture applications, there remain open challenges with training, deployment, and distribution shifts, as identified and discussed below.

\subsection{Data collection}
The process of data collection for training FMs in agricultural applications presents unique challenges due to the inherent variability and diversity in agriculture, influenced by factors such as crop types, growth stages, soil conditions, weather patterns, farming practices, etc. These factors result in a complex and highly diverse dataset, making its collection and standardization a challenging task \citep{lu2020survey, xu2023plant}. Firstly, it is critical to ensure the collected data is accurate, comprehensive, and adequately captures the wide variability found in different farming environments or diverse range of growth phases. However, acquiring high-quality data can be time-consuming, labor-intensive, and expensive, especially considering the need for ground truth labeling in supervised learning scenarios \citep{li2023label} and the variety of growth stages of plants or livestock. To tackle these challenges, generative models \citep{lu2022generative} that create high-fidelity images and label-efficient learning methods to save on tedious labeling costs \citep{li2023label, zhang2023labelbench} show promises. In addition, overcoming the time-consuming and labor-intensive data collection in the field can be achieved by leveraging advancements in autonomous data collection technologies, such as UAVs or UGVs equipped with sensors, as well as remote sensing techniques and the IoT devices \citep{mueller2017robotanist, del2021unmanned, xu2022review}. These technologies automate data-gathering processes and reduce the need for manual intervention.
Secondly, data privacy and ownership pose challenges as farms are typically privately owned. Farmers may be reluctant to share data due to concerns over privacy or potential commercial exploitation. Novel frameworks, such as federated learning \citep{zhang2021survey}, can address these concerns, paving the way for broader access to high-quality, diverse agricultural datasets.
Temporal dynamics represent another significant hurdle. Agricultural processes inherently change over time, influenced by daily fluctuations, seasonal variations, or annual changes. This requires the collection of time-series data (e.g., data across years), introducing another layer of complexity. To handle temporal dynamics, techniques such as time-series analysis \citep{ostrom1990time}, dynamic models based on Long Short-Term Memory (LSTM) \citep{graves2012long}, and data augmentation \citep{lu2022generative} can be employed to better account for the inherent temporal dynamics in agricultural data.

In conclusion, while data collection for training FMs in agriculture is a challenging endeavor, addressing these issues is crucial for developing robust and effective models capable of advancing the state of agricultural technology and practice.


\subsection{Training efficiency} \label{sec:training_fms}
Training FMs for agriculture applications presents significant challenges, particularly with respect to long training times and substantial costs, e.g., requiring thousands of GPU hours and millions of training data. However, innovative solutions are being devised to overcome these challenges. One such approach is the integration of an existing LLM with a relatively lightweight visual prompt generator (VPG) \citep{li2023blip}. Though this approach can reduce the computational costs, the tuning of the VPG component within the VL-LLM still represents a significant computational demand \citep{zhang2023transfer}. A promising solution discussed in the literature involves the transfer of an existing VPG from any existing FMs to the target FMs. This allows for the leveraging of pre-existing model structures and parameters, thereby bypassing the need for extensive training from scratch. The key to optimizing this process lies in maximizing transfer efficiency across different LLM sizes and types. A significant development in this regard is the VPGTrans approach \citep{zhang2023transfer}, a two-stage transfer framework designed to speed up the transfer learning process, which demonstrated that their method can facilitate VPG transfer with a speed-up of more than $10 \times$ and uses only a fraction of the training data compared to building a VPG from scratch.

In summary, while training FMs for agricultural applications indeed presents substantial challenges due to long training times and high costs, advancements like the VPG transfer technique offer promising solutions. These techniques pave the way for more efficient and cost-effective model training processes, thereby making the use of FMs more accessible and practical in the field of agriculture.

\subsection{Distribution shift} \label{sec:df}
The issue of distribution shift presents a significant challenge when applying FMs to agricultural applications. Distribution shift occurs when the data that the model encounters during deployment differs significantly from the data used during its training phase \citep{wiles2022a}. In the context of agriculture, the environmental conditions under which data is collected can vary greatly across different regions and seasons. These variations can include differences in crop types, soil conditions, weather patterns, and farming practices, all of which can cause considerable shifts in data distribution \citep{wiles2022a}. For example, \cite{williams2023leaf} demonstrated that the direct application of FMs in a zero-shot manner for leaf segmentation tasks resulted in unsatisfactory performance, likely due to potential distribution shifts \citep{chawla2021quantifying}. 

To combat the challenges posed by distribution shifts when training and deploying FMs, researchers have been continuously investigating these issues \citep{chawla2021quantifying, cvejoski2022future, bommasani2021opportunities}. For instance, \cite{cvejoski2022future} demonstrated that LLMs may experience performance degradation of up to approximately 88\% when predicting the popularity of future posts from subreddits whose topic distribution evolves over time. To mitigate these distribution shift issues, they introduced neural variational dynamic topic models and attention mechanisms. These techniques infer temporal language model representations for regression tasks, resulting in only about 40\% performance drops in the worst-case scenarios and a mere 2\% in the best cases. Furthermore, \cite{wang2023sam} demonstrated that the integration of techniques such as multi-task learning, continual learning, and distillation with VFMs like SAM and CLIP can significantly alleviate distribution shift issues. 
These techniques have been shown to significantly mitigate performance degradation caused by distribution shifts, thus demonstrating promising prospects for the application of FMs in the diverse and dynamic field of agriculture.

\subsection{Other challenges in real-world deployment}
Deploying FMs for agricultural applications presents a unique set of challenges, particularly regarding inference speed and model size. Firstly, the inherent complexity and size of these models can pose substantial computational demands. Large-scale FMs often require significant processing power and memory for both training and inference, which may not always be readily available in on-field agricultural settings \citep{bommasani2021opportunities}. This can limit the deployment of these models in real-time applications, such as monitoring crops or automating irrigation systems. Secondly, inference speed is crucial for many agricultural tasks that require immediate action based on the model's outputs \citep{bommasani2021opportunities}. For instance, in precision agriculture, time-sensitive tasks like pest detection or predicting weather impacts on crop health demand quick and accurate responses. However, due to the size and complexity of FMs, achieving high-speed inference can be a challenge.

To overcome these hurdles, future research and development efforts need to focus on model optimization techniques  \citep{zhong2023transformer, zhu2023survey}, such as model compression (e.g., pruning and quantization \citep{cheng2017survey, choudhary2020comprehensive}) and efficient network design \citep{wan2022efficiently}, which could potentially reduce the model size without compromising performance. Actually, there are already some attempts to deploy FMs. For instance, FQ-ViT \citep{lin2022fq} introduces a fully quantized vision transformer that simplifies inference by utilizing only 1/4 of the memory with minimal accuracy degradation (approximately 1\%). Additionally, SqueezeLLM \citep{kim2023squeezellm} demonstrates that utilizing only 3/16 of the parameters significantly reduces the perplexity gap by up to 2.1 times and achieves up to 2.3 times speedup compared to baseline models.

Additionally, the utilization of edge computing solutions could expedite the inference process by enabling data processing closer to its source, consequently enhancing the speed and efficiency of FMs in agricultural applications \citep{chen2019deep}. For example, in agricultural applications, edge devices like drones or field sensors can analyze data locally, enabling real-time decision-making without relying on distant servers. Recently, MLC LLM \citep{mlcllm2023} offers a universal solution, allowing any language model to be deployed directly across a diverse range of hardware backends and native applications. This solution also provides a productive framework for individuals to further optimize model performance for their specific use cases. On the other hand, MobileSAM \citep{mobile_sam} and MobileSAMv2 \citep{zhang2023mobilesamv2} make SAM \citep{kirillov2023segment} mobile-friendly by replacing the heavyweight image encoder with a lightweight one. This lightweight SAM is more than 60 times smaller yet performs on par with the original SAM \citep{kirillov2023segment}.

\subsection{Building capacity for FM adoption in agriculture}
Effective utilization of FMs in agriculture requires specialized skills and multi-discipline collaborations \citep{mai2023opportunities}. However, there are several challenges in the skill transfer of applying FMs in agricultural domains, including domain adaptation techniques due to agricultural researchers' potential lack of familiarity with FM applications, understanding unique characteristics of agricultural data such as high variability, and interpreting the complex nature of FMs.

Addressing the skill gap and providing targeted training is crucial for empowering agricultural researchers and users to leverage FMs optimally. Investing in training and capacity-building initiatives, such as workshops, webinars, specialized training programs focusing on FM utilization in agriculture, and educational resources such as online tutorials and case studies, is essential for enabling effective FM utilization in agriculture, driving innovation, and addressing agricultural challenges.

\section{Conclusion}
\label{sec:sum}

In conclusion, this study has investigated the potential of FMs in the field of smart agriculture. These models hold the promise of transforming the agricultural sector by reducing reliance on extensive labeled datasets and enhancing generalization and effectiveness. By categorizing and reviewing various types of LLMs and FMs (Section~\ref{sec:fms}), their applicability in smart crop management, smart plant breeding, smart livestock farming, precision aquaculture farming, and agricultural robotics are revealed (Section~\ref{sec:afms}). However, it is crucial to address challenges and risks such as data collection, model training, distribution shift, real-world deployment, and research gaps before real-world implementation (Section~\ref{sec:dis}). Further exploration and utilization of LLMs and FMs have the potential to enhance productivity, sustainability, and decision-making in agriculture.

In the era of LLMs and FMs, all agricultural fields successfully explored using traditional ML/DL approaches, as discussed in Section~\ref{sec:afms} and beyond, are ready for exploration by employing FMs. Integrating LLMs and FMs into the existing framework is also a promising direction to improve efficiency and effectiveness, as evident in the studies by \cite{stella2023can, tzachor2023large}. These integrations can leverage the strengths of both frameworks, potentially leading to more robust and adaptable agricultural systems. Furthermore, the development of large-scale agricultural datasets would be an approaching direction to accelerate the development and application of LLMs and FMs in the agriculture field \citep{lu2020survey}. Such datasets can facilitate more comprehensive training of models, enabling them to capture the nuances and complexities of agricultural processes more accurately. Additionally, there is a growing need for interdisciplinary collaboration between experts in agriculture, computer science, and data science. By fostering collaboration across these domains, innovative solutions can be developed to address the unique challenges and opportunities in agriculture using LLMs and FMs.

Ultimately, this study aims to inspire and guide researchers to unlock the untapped potential of LLMs and FMs in smart agriculture and serves as a stepping stone toward developing smart agriculture with next-level intelligence and more sophisticated capabilities.

\section*{Authorship Contribution}
\textbf{Jiajia Li}: Conceptualization, Investigation, Methodology, Software, Writing – original draft;
\textbf{Mingle Xu}: Formal Analysis, Writing - review \& editing; 
\textbf{Lirong Xiang}: Formal Analysis, Writing - review \& editing;
\textbf{Dong Chen}: Conceptualization, Investigation, Writing – original draft; 
\textbf{Weichao Zhuang}: Writing - review \& editing;
\textbf{Xunyuan Yin}: Writing - review \& editing; 
\textbf{Zhaojian Li}: Supervision, Resources, Writing - review \& editing.


\typeout{}
\bibliography{ref}
\end{document}